\documentclass[10pt,twocolumn,letterpaper]{article}

\usepackage{iccv}
\usepackage{times}
\usepackage{epsfig}
\usepackage{graphicx}
\usepackage{amsmath}
\usepackage{amssymb}
\usepackage{svg}
\usepackage{adjustbox}

\usepackage{booktabs}
\usepackage{subcaption}
\usepackage{float}
\usepackage{multirow}
\usepackage{multicol}
\usepackage{makecell}
\usepackage{pifont}
\usepackage{bbm}

\usepackage[accsupp]{axessibility}
\usepackage{amsmath}  

\usepackage[
breaklinks=true,
bookmarks=true,
colorlinks=true,
pdfpagemode=UseOutlines, 
bookmarksnumbered=true   
]{hyperref}

\iccvfinalcopy 

\def\iccvPaperID{****} 

\usepackage[bottom]{footmisc}
\usepackage{pdfcomment}
\usepackage{xcolor}

\begin{document}

\title{SegChange-R1: LLM-Augmented Remote Sensing Change Detection}

\author{Fei Zhou$^{1,2}$\\
		$^1$
        Neusoft Institute Guangdong, China $^2$Airace Technology Co.,Ltd., China\\
		{\tt\small zhoufei21@s.nuit.edu.cn}
	}

\maketitle


\begin{abstract}

Remote sensing change detection is used in urban planning, terrain analysis, and environmental monitoring by analyzing feature changes in the same area over time. In this paper, we propose a large language model (LLM) augmented inference approach (SegChange-R1), which enhances the detection capability by integrating textual descriptive information and guides the model to focus on relevant change regions, accelerating convergence. We designed a linear attention-based spatial transformation module (BEV) to address modal misalignment by unifying features from different times into a BEV space. Furthermore, we introduce DVCD, a novel dataset for building change detection from UAV viewpoints. Experiments on four widely-used datasets demonstrate significant improvements over existing method The code and pre-trained models are available in \url{https://github.com/Yu-Zhouz/SegChange-R1}.

\end{abstract}


\section{Introduction}

Change detection (CD) in remote sensing analyzes images of the same area acquired at different times to identify surface feature changes~\cite{bai2023deep,khelifi2020deep}. CD tasks are used for urban expansion monitoring~\cite{xu2016gis,zhang2023aerial}, disaster assessment~\cite{atzberger2013advances,tuia2012multitemporal}, land use and land cover change analysis~\cite{coppin2004review,singh1989review,zhu2014continuous}, and military reconnaissance~\cite{gong2015change,jiao2020change}. However, CD faces challenges due to various factors affecting remote sensing images.

First, images from different time points often exhibit lighting and seasonal variations, causing spectral differences in the same objects~\cite{bruzzone2000automatic,liu2015sequential}. Second, inconsistent resolution in multi-source data can affect the accuracy of change extraction~\cite{bovoloframework}. Sensor noise and atmospheric interference also introduce image noise, complicating change modeling~\cite{bai2023deep,wang2022aerial}. Image registration errors are another critical issue; even after preprocessing, minor misalignments can lead to incorrect change judgments~\cite{bovoloframework,radke2005image}.

Recently, convolutional neural networks (CNNs) such as Fully Convolutional Early Fusion (FC-EF) and FC-Siam-diff have enhanced feature consistency in dual-temporal images using twin structures, improving detection performance. Transformer-based methods, such as BIT~\cite{chen2021remote} and ChangeFormer~\cite{bandara2022transformer}, use self-attention to model long-range dependencies, improving multi-scale change modeling. ChangeMamba~\cite{lee2024changemamba} is based on the Mamba architecture and utilizes a state space model to process long sequences of remote sensing data, improving modeling efficiency. However, most methods are limited to visual feature extraction and lack semantic understanding, affecting the accuracy and convergence speed of change features. Recently, Bird's Eye View (BEV) space modeling has been introduced to unify perspective representation, aiming to address this issue. Combining contextual information, such as text descriptions or geographic tags, to focus the model on regions of interest remains an area for exploration~\cite{li2020comprehensive}.

Recently, the rapid development of LLMs has brought new opportunities to this field. We propose SegChange-R1, which uses LLMs to combine textual descriptions with remote sensing images. This guides the model to focus on regions of interest, improving the detection of significant changes between time phases. As shown in Fig. 2, we designed a spatial transformation module (BEV) based on linear attention. This module unifies features from different temporal perspectives into a shared BEV space, addressing modal mismatch in change detection. Compared to transformers, the linear attention-based architecture enhances feature expression by modeling global dependencies, enabling linear-time training, efficient spatial dependency modeling, and rapid convergence. In summary, the main contributions of our work are as follows:

\begin{itemize}
    \setlength\itemsep{0.1em}
    \item We developed SegChange-R1, a novel semantic-guided remote sensing change detector that uses a large language model to generate accurate location masks by integrating textual descriptions of two images. 
    
    \item We designed a linear attention-based spatial transformation module (BEV) to address modal mismatch in change detection.
    
    \item We introduce DVSC, a drone-view building change detection dataset containing 13,800 image pairs of building changes in diverse urban and rural scenarios.
\end{itemize}


\section{Related Work} 

\paragraph{Deep Learning Based methods.}
Remote sensing change detection has evolved from traditional methods to deep learning methods. Early change detection methods primarily relied on pixel-level difference analysis, including image difference methods, ratio methods, change vector analysis, and principal component analysis~\cite{bai2023deep,khelifi2020deep,singh1989review}. While these traditional methods are computationally simple, they are often influenced by various factors such as changes in lighting, seasonal variations, sensor noise, and atmospheric conditions, leading to high false alarm rates and making it difficult to accurately identify genuine changes in land cover.

With the development of deep learning, convolutional neural networks (CNNs) have become the mainstream paradigm for change detection. Early deep learning methods such as FC-EF (Fully Convolutional Early Fusion) employed an early fusion strategy, concatenating dual-phase images at the input layer~\cite{daudt2018fully}. Subsequently, twin network architectures such as FC-Siam-diff (Fully Convolutional Siamese Difference) and FC-Siam-conc were proposed, which perform parallel processing of dual-phase images using feature extractors with shared weights, followed by feature fusion through difference or concatenation operations~\cite{daudt2018urban,marin2019recipe1m+}. These methods ensure feature consistency through parameter sharing, enhancing the accuracy and robustness of change detection.

To improve detection performance, researchers have explored complex network architectures and attention mechanisms. STANet~\cite{chen2020spatial} introduced a spatio-temporal attention mechanism, enhancing the model's ability to focus on changing regions through spatial attention and channel attention modules. DT-CDSCN~\cite{feng2020dtcdscn} proposed a dual-task constrained deep twin convolutional network, improving change detection performance through auxiliary training with a semantic segmentation task. IFN~\cite{zhang2023changeclicp} designed an interactive feature fusion network, achieving more precise change modeling through multi-level feature interaction. In recent years, attention-based methods such as SNUNet~\cite{fang2021snunet} and FCCDN~\cite{chen2022fccdn} have further advanced this field, enhancing feature representation and fusion capabilities through the design of specialized attention modules.

\vspace{-10pt}
\paragraph{Transformer-based methods.}
To better model long-range dependencies and global context, Transformer-based change detection methods have emerged. BIT (Binary Change Detection with Transformers)~\cite{chen2021remote} was the first to introduce the Transformer architecture into change detection tasks, using self-attention mechanisms to capture global contextual information, achieving significant performance improvements across multiple benchmark datasets. ChangeFormer~\cite{bandara2022transformer} further improved the Transformer structure by designing specialized change-aware attention modules and hierarchical feature fusion strategies. SwinSUNet~\cite{wang2022unetformer} combines the hierarchical feature representation capabilities of the Swin Transformer with a sliding window mechanism, demonstrating outstanding performance when handling multi-scale changes.  

While Transformers excel at modeling global dependencies, their $O(n^2)$ complexity becomes a bottleneck when processing high-resolution remote sensing images. To address this issue, researchers have begun exploring more efficient attention mechanisms and alternative architectures. ChangeMamba~\cite{lee2024changemamba} designs a change detection architecture with linear complexity based on a state space model (SSM), achieving efficient long sequence modeling through Mamba's selective scanning mechanism. Performer~\cite{choromanski2020rethinking} achieves linear-complexity attention computation through random feature mapping, while Linformer~\cite{wang2020linformer} reduces the dimension of the attention matrix via low-rank decomposition. These methods reduce computational costs while maintaining performance, making them suitable for high-resolution remote sensing images.

\vspace{-10pt}
\paragraph{Methods for multimodal fusion.}
Existing change detection methods emphasize visual representation learning, neglecting the potential of multimodal data. CLIP (Contrastive Language-Image Pre-training)~\cite{radford2021learning} successfully demonstrated the effectiveness of visual-language pre-training across various visual tasks, laying the foundation for multimodal applications in the field of remote sensing. Subsequent works such as FLAVA~\cite{singh2022flava} and ALIGN~\cite{jia2021scaling} further explored the possibilities of large-scale multimodal pre-training. Inspired by this, some researchers began to incorporate language information into remote sensing change detection tasks, exploring how text descriptions can guide models to focus on specific types of changes~\cite{li2020comprehensive,liang2020deeplearning,zhang2023changeclicp,liu2023promptcc}. Recently, the rapid development of LLMs has brought new opportunities to remote sensing change detection. Multimodal large language models such as Generative Pre-trained
Transformer 4 with Vision (GPT-4V)~\cite{openai2023gpt}, LLaVA~\cite{liu2023visual}, and InstructBLIP~\cite{dai2023instructblip} have demonstrated strong visual understanding and reasoning capabilities. In the field of remote sensing, LLMs are increasingly being applied to tasks such as image description generation, scene understanding, and object detection~\cite{cha2024vision,yuan2024change,sumbul2022multimodal}. In particular, LLMs' capabilities in spatial reasoning and regional localization offer new possibilities for remote sensing change detection. By combining visual features with natural language descriptions, LLMs can better understand the semantics of changes, guiding models to focus on regions of interest.  

However, multimodal fusion in remote sensing change detection remains limited. Most methods still rely on visual feature comparisons, lacking a deep understanding of semantic changes. Effectively integrating LLM reasoning into change detection and achieving deep fusion of visual features and language remains a critical issue.

\vspace{-10pt}
\paragraph{Spatial Alignment and Efficient Architectures.}

Spatial alignment and efficient architecture are key technical challenges in remote sensing change detection. Due to various interference factors during remote sensing image acquisition, such as sensor position differences, changes in shooting angles, and atmospheric conditions, images from different time phases often exhibit minor but non-negligible spatial misalignment issues~\cite{bovoloframework2,radke2005image}. This misalignment can lead to erroneous change detection, even at the sub-pixel level, particularly in edge regions and for small targets.  

Bird's Eye View (BEV) representation, as a unified spatial representation method, has been widely applied in autonomous driving and 3D object detection fields~\cite{fang2023bevheight,philion2020lift,zhou2022cross}. BEV's advantage lies in unifying data from different perspectives and sensors into a single spatial coordinate system, addressing perspective changes and spatial misalignment. LSS (Lift, Splat, Shoot)~\cite{philion2020lift} converts perspective view features into BEV representation through depth estimation, while BEVFormer~\cite{li2022bevformer} further incorporates temporal information modeling. Recently, some researchers have begun to incorporate BEV representations into remote sensing change detection tasks to address registration errors between multi-temporal images~\cite{li2023bev,chen2021self}. By mapping image features from different temporal phases into a unified BEV space, the impact of spatial misalignment on change detection can be reduced.


\begin{figure*}[ht]
	\centering
	\includegraphics[width=\linewidth]{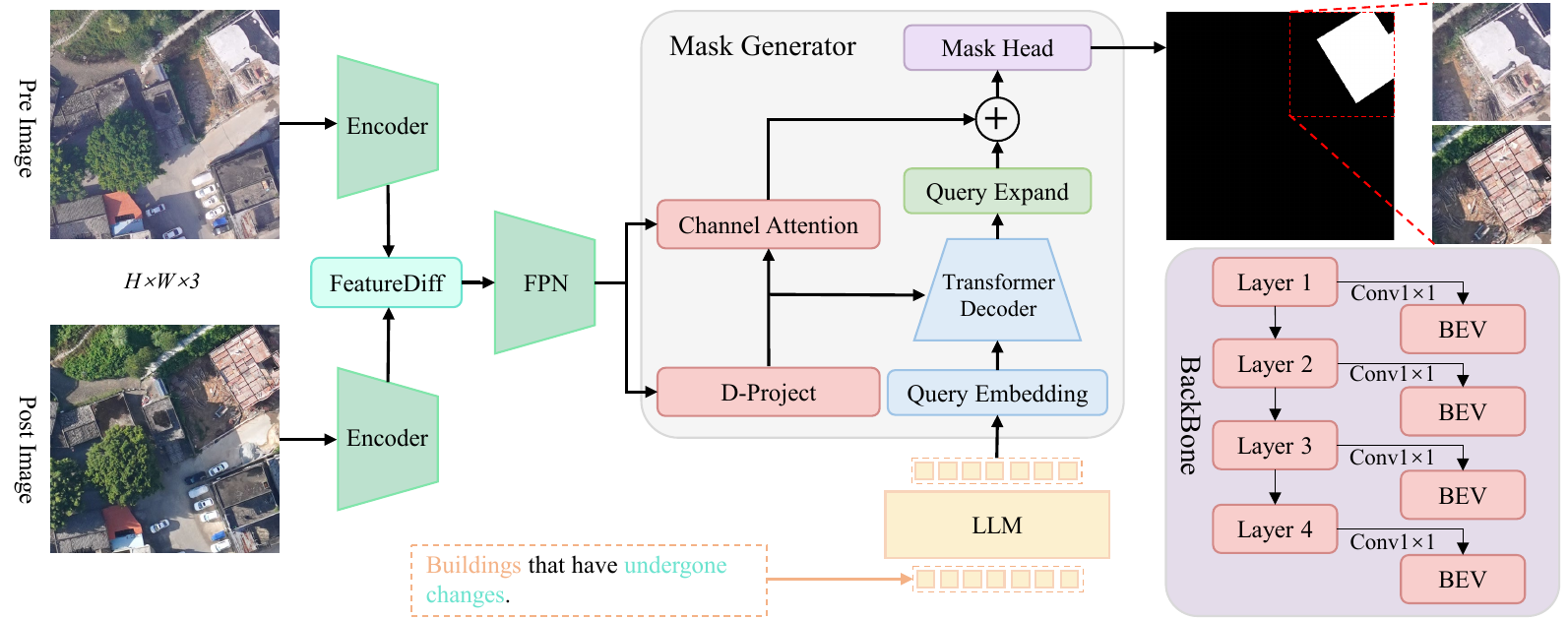}
	\caption{\textbf{An overview of the SegChange-R1 Architecture}. Given two images before and after a change, they are passed through a shared BackBone to extract multi-scale features, which are then input into the difference module for feature difference extraction. The features are then sent to the Feature Pyramid Network (FPN) layer for fusion. Finally, the resulting feature maps and text semantic features are input into the Mask Head to predict the feature map of the changed area.} 
	\label{fig:structure}
	\vspace{-5pt}
\end{figure*}

\section{Proposed Method}

We propose SegChange-R1, a geospatial pixel inference baseline. As shown in Fig.~\ref{fig:structure}, our architecture includes a pre-trained text-image encoder, a BEV space transformer, and a masked decoder.

\subsection{Encoder}

\paragraph{Vision Encoder}

Remote sensing images exhibit significant scale variation, from sub-meter objects to kilometer-scale structures, challenging the model's multi-scale modeling capabilities~\cite{zhou2022cross}. Additionally, the dense distribution of small targets in high-resolution images requires the model to retain spatial detail during feature extraction~\cite{choromanski2020rethinking}. However, mainstream Vision Transformer encoders (such as CLIP~\cite{bandara2022transformer} and Segment Anything Model (SAM)~\cite{fang2021snunet,singh2022flava}) have limitations: their fixed window mechanism and aggressive downsampling can lead to the loss of small-scale target information, limiting perception in complex remote sensing scenarios. To address these issues, we adopted the Swin Transformer~\cite{atzberger2013advances}, which uses a sliding window to model local attention, enhancing the capture of fine-grained features while maintaining efficiency. Based on this, we constructed a progressive multi-scale feature extraction framework, generating feature maps at resolutions of 1/4, 1/8, 1/16 and 1/32 of the original input image, denoted as $v_h\in[1, 4]$, thereby balancing spatial resolution and semantic abstraction at different levels. Additionally, our implementation supports multiple backbones (including ResNet50~\cite{he2016deep}, Swin-Transformer~\cite{liu2021swin}, and HGNetv2~\cite{zhao2024detr}) to accommodate diverse computational and accuracy requirements.

\vspace{-10pt}
\paragraph{Text Encoder}
Text prompts are crucial in remote sensing change detection, providing semantic information about changes and guiding the model to focus on specific features. Vision-language models have demonstrated that incorporating text semantics effectively enhances visual understanding~\cite{radford2021learning}. ChangeCLIP~\cite{zhang2023changeclicp} leverages comprehensive text semantics from remote sensing images to enhance visual models' perception of changes, achieving strong performance on benchmark datasets. Additionally, studies such as PromptCC~\cite{liu2023promptcc} validate the effectiveness of using LLMs for text prompt encoding; multi-prompt learning improves language generation accuracy.

To bridge the semantic gap between text and visual features, we introduce a text encoder based on a pre-trained Large Language Model (LLM), Microsoft/Phi-1.5. This encoder converts text descriptions into semantically rich embeddings, integrated with the visual encoder through deep feature fusion. This enables the model to focus on specific land cover changes guided by semantic constraints from text. Additionally, we implemented a dynamic sequence length control, enabling text embeddings to adapt to downstream tasks.

\subsection{BEV Space Converter}
Modal mismatch is a major challenge in remote sensing change detection. To address this, we propose a BEV space transformation module in our SegChange-R1 framework. This module addresses modal mismatch when processing data from different times and is based on a linear attention mechanism for efficient feature transformation. The module transforms features from different times into a shared BEV space, enabling effective comparison and change analysis.

The BEV Space Converter takes features from multiple times as input, projecting them into a latent space using linear transformations. Mathematically, this can be represented as follows: 
\begin{align}
\mathbf{z}_t = W_z \mathbf{x}_t + \mathbf{b}_z  
\end{align}
here, $( \mathbf{x}_t)$ represents the input features from time phase $t$, $W_z$ is the learnable weight matrix, and $mathbf{b}_z$ is the bias term. The transformed features $mathbf{z}_t$ are then used to compute attention scores.  

The attention scores are calculated using the linear attention mechanism. For each position \( i \) in the feature map, the attention score $a_{ij}$ with respect to position $j$ is computed as:  
\begin{align}
A_{ij} = \mathbf{w}_a^\top \text{ReLU}(\mathbf{W}_{a 1} \mathbf{z}_{i} + \mathbf{W}_{a 2} \mathbf{z}_{j})  
\end{align}
where $\mathbf{w}_a$,  $\mathbf{W}_{a 1}$, and $\mathbf{W}_{a 2}$ are learnable parameters. These attention scores are then normalized using the softmax function to obtain the final attention weights.  

Using these attention weights, the features are aggregated into a unified BEV representation, allowing the model to address modality misalignment and capture changes between different times.  

The BEV Space Converter enhances change detection and provides a robust, interpretable feature representation.


\begin{table*}[htbp]
	\centering
	\begin{tabular}{lcccccccc}
		\specialrule{0.5pt}{0pt}{0pt} 
		\textbf{Dataset} & \textbf{Size} & \textbf{Instruct} & \textbf{Train} & \textbf{Val} & \textbf{Test} & \textbf{Time Span} & \textbf{Spatial Resolution} & \textbf{Coverage Area} \\
		\hline
		WHU-CD & 512×512 & \(\times\) & 5947 & 743 & 744 & 2012-2016 & 0.075m & New Zealand \\
		DSIFN-CD & 512×512 & \(\times\) & 14400 & 1360 & 192 & 2015-2018 & 2m & Six Cities in China \\
		CDD & 256×256 & \(\times\) & 10000 & 2998 & 3000 & 2006-2019 & 3-100cm & Global \\
		DVCD & 512×512 & \checkmark & 11066 & 1383 & 1384 & 2022-2024 & 0.1m & Guangdong Towns \\
		\specialrule{0.5pt}{0pt}{0pt} 
	\end{tabular}
	\caption{Datasets used for building change detection. This table summarizes the characteristics of three publicly available benchmark datasets, along with details of the newly proposed dataset.}
	\label{table:dataset}
\end{table*}

\subsection{Mask Decoder}

\begin{figure}[ht]
	\centering
	\includegraphics[width=1.0\linewidth]{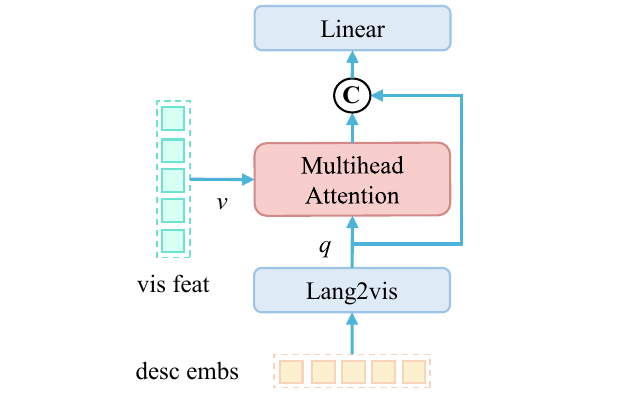}
	\caption{\textbf{D-Project}. By embedding text into visual features, aligning features through cross-attention, and finally obtaining a refined prediction map through decoder interaction and mask prediction.
	}
	\vspace{-10pt} 
	\label{fig:D-Project}
\end{figure}

The mask decoder generates precise segmentation masks by integrating multi-scale visual features with semantic guidance from text. Inspired by language-guided segmentation~\cite{bai2023deep}, our architecture uses a hierarchical design that combines cross-modal feature fusion with transformer-based spatial reasoning. As shown in Fig.~\ref{fig:D-Project}, the decoder includes: a description projector (D-Projector) that bridges the semantic gap between language and vision, a transformer decoder that performs spatial reasoning through learnable queries, and a mask prediction head that generates the segmentation output.

The D-Projector, shown in Fig.~\ref{fig:D-Project}, first aggregates temporal text embeddings and projects them into the visual feature space. Then, it uses cross-attention to achieve fine-grained alignment between text semantics and spatial visual features. Subsequently, the transformer decoder interacts with the fused visual features through self- and cross-attention using learnable queries, enabling the model to capture long-range spatial dependencies and contextual associations for accurate segmentation. The final mask prediction head employs a multi-scale convolutional architecture with channel attention to refine spatial correlations and generate high-quality segmentation masks. This design enables the model to handle complex spatial reasoning while maintaining efficiency, as validated by our experiments.


\begin{table*}[htbp]
	\centering
	\adjustbox{width=\textwidth,center}{
		\begin{tabular}{lccccccccccccc}
			\specialrule{0.5pt}{0pt}{0pt}
			& & \multicolumn{3}{c}{\textbf{WHU-CD}} & \multicolumn{3}{c}{\textbf{DSIFN-CD}} & \multicolumn{3}{c}{\textbf{CDD}} & 
			\multicolumn{3}{c}{\textbf{DVCD}} \\
			\textbf{Method} & \textbf{Instruct} & \textbf{F1$\uparrow$} & \textbf{IoU$\uparrow$} & \textbf{OA$\uparrow$} & \textbf{F1$\uparrow$} & \textbf{IoU$\uparrow$} & \textbf{OA$\uparrow$} & \textbf{F1$\uparrow$} & \textbf{IoU$\uparrow$} & \textbf{OA$\uparrow$} & \textbf{F1$\uparrow$} & \textbf{IoU$\uparrow$} & \textbf{OA$\uparrow$} \\
			\hline
			FC-Siam-conc & $\times$ & 0.798 & 0.665 & 98.50 & 0.579 & 0.426 & 89.60 & 0.751 & 0.601 & 94.90 & - & - & - \\
			STANet & $\times$ & 0.823 & 0.700 & 98.50 & 0.645 & 0.478 & 88.50 & 0.841 & 0.722 & 96.10 & - & - & - \\
			DT-CDSCN & $\times$ & 0.914 & 0.842 & 99.30 & 0.706 & 0.545 & 82.90 & 0.921 & 0.853 & 98.20 & - & - & - \\
			SNUNet & $\times$ & 0.835 & 0.717 & 98.71 & 0.662 & 0.495 & 87.34 & 0.962 & - & - & - & - & - \\
			FCCDN & $\times$ & 0.937 & - & - & - & - & - & 93.73 & 88.20 & - & - & - & - \\
			BIT & $\times$ & 0.839 & 0.724 & 98.75 & 0.693 & 0.529 & 89.41 & - & - & - & - & - & - \\
			ChangeFormer & $\times$ & 0.886 & 0.795 & 99.12 & \underline{0.947} & \underline{0.887} & \underline{93.20} & 0.946 & 0.898 & 98.70 & - & - & - \\
			SwinSUNet & $\times$ & 0.938 & - & 99.40 & - & - & - & - & - & - & 0.896 & 0.801 & 92.37 \\
			ChangeMamba & $\times$ & 0.942 & 0.890 & 93.98 & - & - & - & - & - & - & 0.908 & 0.825 & 96.56 \\
			ChangeCLIP & $\checkmark$ & \textbf{0.982} & \underline{0.915} & \underline{99.52} & - & - & - & \underline{0.979} & \underline{0.975} & \underline{99.48} & \underline{0.913} & \underline{0.858} & \underline{98.20} \\
			SegChange-R1 & $\checkmark$ & \underline{0.968} & \textbf{0.926} & \textbf{99.60} & \textbf{0.972} & \textbf{0.921} & \textbf{96.73} & \textbf{0.988} & \textbf{0.969} & \textbf{99.61} & \textbf{0.919} & \textbf{0.867} & \textbf{98.62} \\
			\specialrule{0.5pt}{0pt}{0pt}
		\end{tabular}
	}
	\caption{Quantitative comparison of SegChange-R1 against state-of-the-art change detection (CD) methods. Best and second-best results are indicated in bold and underlined, respectively. SegChange-R1 demonstrates superior performance on the DSIFN-CD, CDD, and DVCD datasets.}
	\label{table:CD_Results}
\end{table*}


\section{Experimental Setup and Results} 

\paragraph{Datasets}
The rapid development of drone aerial photography, with its flexible deployment, high resolution, and multi-angle imaging, has opened new possibilities for urban environment monitoring. However, most change detection datasets are based on traditional remote sensing platforms and lack semantic-level modeling. They are constrained by low resolution and fixed angles, making it difficult to capture diverse building details in complex urban environments. This motivates enhancing the model's understanding of semantic changes and exploring the role of text in building change detection. 

We constructed DVCD (Drone View Change Detection), a new drone-based change detection dataset. This dataset consists of drone orthophotos collected from urban areas in Guangdong Province, covering building changes from 2022 to 2024. It reflects typical changes such as new construction, demolition, and expansion during rapid urban development. We introduced fine-grained textual descriptions to guide the model to focus on semantic change features, enhancing its ability to recognize building evolution in complex scenarios. The DVCD dataset comprises 12,833 image pairs, divided into a training set (11,066 pairs), a validation set (1,383 pairs), and a test set (1,384 pairs).

To further validate the algorithm's generalization ability, this study also conducted comparative experiments on three publicly available building change detection benchmark datasets, with details as shown in Table.~\ref{table:dataset}. The WHU-CD~\cite{daudt2018fully} dataset is specifically designed for building extraction and change detection tasks; the DSIFN-CD~\cite{zhang2020deeply} dataset consists of large-scale high-resolution dual-time-phase images covering six cities in China; The CDD~\cite{lebedev2018change} dataset provides diverse image pairs with varying resolutions and seasonal changes. These datasets have unique characteristics in image size, sample quantity, temporal span, spatial variation, and geographical coverage, providing a foundation for evaluating algorithm performance.

\vspace{-10pt}
\paragraph{Experimental Setup}
All experiments were conducted on servers with NVIDIA A800 Graphics Processing Units (GPUs). Training used the AdamW optimizer with a learning rate of $10^{-4}$, a backbone network learning rate of $10^{-5}$, and a weight decay of $10^{-4}$. Training used 128 epochs with a batch size of 16 and a StepLR scheduler, reducing the learning rate by a factor of 0.1 every 20 epochs. During testing, the batch size was 1, and the change detection threshold was 0.5. Model performance was evaluated using F1 Score (F1-Score; harmonic mean of precision and recall), Intersection over Union (IoU), and Overall Accuracy (OA).

\vspace{-10pt}
\paragraph{Results}
We compare our method with recent state-of-the-art change detection methods on four datasets. As shown in Table.~\ref{table:CD_Results}, our proposed SegChange-R1 demonstrated better performance in all benchmark tests. Among the existing methods, CNN-based models (e.g., FC-Siam-conc and~\cite{fang2021snunet}) have limited effectiveness due to their limited global context modeling capabilities. Transformer-based approaches (e.g., ChangeFormer~\cite{bandara2022transformer} and SwinSUNet~\cite{wang2022unetformer}) improve performance by modeling long-range dependencies through a self-attentive mechanism. Notably, ChangeCLIP~\cite{zhang2023changeclicp} introduces instruction-guided learning and utilizes visual language pre-training to achieve good results on WHU-CD (F1: 0.982) and CDD (F1: 0.979), highlighting the advantages of incorporating semantic guidance. In contrast, SegChange-R1 utilizes text-guided semantic understanding, achieving the highest performance on DSIFN-CD (F1: 0.972) and CDD (F1: 0.988), and achieves the highest accuracy on the drone-view DVCD dataset, highlighting its effectiveness in complex scenarios.

For the DVCD dataset evaluation, we selected advanced open-source methods and followed their original training configurations for fair comparison. Notably, SegChange-R1 demonstrated high training efficiency, converging in 64 epochs and outperforming all competitors.


\begin{table}[htbp]
	\centering
	\adjustbox{width=0.5\textwidth,center}{
		\begin{tabular}{lcccccc}
			\specialrule{0.5pt}{0pt}{0pt} 
			\textbf{Backbone} & \textbf{F1} & \textbf{IoU} & \textbf{OA} & \textbf{Parameters(M)} & \textbf{FLOPs(G)} & \textbf{Time(s)} \\
			\hline
			ResNet-50 & 87.34 & 78.82 & 94.15 & 52.3 & 45.2 & 28.5 \\
			HGNetV2 & 89.12 & 83.45 & 95.23 & 48.7 & 38.9 & 24.2 \\
			\textbf{Swin-Base} & \textbf{91.78} & \textbf{86.70} & \textbf{98.62} & \textbf{88.9} & \textbf{67.4} & \textbf{35.8} \\
			\specialrule{0.5pt}{0pt}{0pt} 
		\end{tabular}
	}
	\caption{Performance comparison of backbone networks. The Swin Transformer backbone achieved the highest performance.}
	\label{table:backbone}
\end{table}


\begin{table}[htbp]
	\centering
	\adjustbox{width=0.5\textwidth,center}{
		\begin{tabular}{lccccc}
			\specialrule{0.5pt}{0pt}{0pt} 
			\textbf{Text Configuration} & \textbf{F1} & \textbf{IoU} & \textbf{OA} & \textbf{Description} \\
			\hline
			No Text & 87.23 & 77.45 & 93.67 & Pure visual features \\
			Static Prompts & 88.94 & 79.82 & 94.23 & Fixed text templates \\
			Dynamic Descriptions & 90.15 & 82.31 & 95.04 & Context-aware text \\
			\textbf{LLM-Enhanced} & \textbf{91.78} & \textbf{86.70} & \textbf{98.62} & \textbf{Full text reasoning} \\
			\specialrule{0.5pt}{0pt}{0pt} 
		\end{tabular}
	}
	\caption{Performance comparison of different prompt configurations for [Task/Model].}
	\label{table:Prompt}
\end{table}


\begin{table*}[htbp]
	\centering
	\adjustbox{width=\textwidth,center}{
		\begin{tabular}{lccccccccc}
			\specialrule{0.5pt}{0pt}{0pt} 
			\textbf{Text} & \textbf{F1} & \textbf{IoU} & \textbf{OA} & \textbf{Spatial} & \textbf{Edge} & \textbf{Parameters(M)} & \textbf{FLOPs(G)} & \textbf{Time(s)} \\
			\textbf{Configuration} & & & & \textbf{Consistency} & \textbf{Accuracy} & & & \\
			\hline
			No BEV & 89.12 & 81.23 & 94.87 & 0.832 & 0.791 & - & - & - \\
			Transformer & 91.34 & 83.89 & 95.87 & 0.886 & 0.867 & 139.99 & 4.28 & 1.19 \\
			\textbf{Linear Attention} & \textbf{91.78} &\textbf{86.70} & \textbf{98.62} & \textbf{0.912} & \textbf{0.869} & \textbf{4.29} & \textbf{0.13} & \textbf{0.85} \\
			\specialrule{0.5pt}{0pt}{0pt} 
		\end{tabular}
	}
	\caption{Effectiveness of the bird's-eye view (BEV) module in improving [Task/Metric].}
	\label{table:bev}
\end{table*}

\section{Ablation Study}

To validate the design and quantify the contribution of each component, we conducted an ablation study on the DVCD dataset. This analysis provided insights for architecture selection and demonstrated the necessity of each module for optimal change detection.

\vspace{-10pt}
\paragraph{Backbone}
We investigated three representative backbone architectures to understand their fundamental capabilities in capturing multi-scale spatial features, which are critical for change detection tasks, as shown in Table.~\ref{table:backbone}. The analysis revealed the impact of different computational paradigms on change detection effectiveness.

While ResNet-50 provides a baseline, its CNN architecture limits global context modeling. HGNetV2 adopts a lightweight architecture, balancing computational efficiency and representation. The Swin Transformer demonstrates strong performance through its hierarchical self-attention, suited to the multi-scale characteristics of change detection. This enables the capture of both fine-grained local changes and global context, resulting in more coherent change detection.

\vspace{-10pt}
\paragraph{Text Guidance Strategy Analysis}
We systematically evaluate the gradual integration of textual semantic information to understand its role in enhancing change detection through multimodal reasoning capabilities, as shown in Table.~\ref{table:Prompt}. The baseline without text relies solely on visual features, limiting the model's ability to integrate semantic context and domain knowledge. This limitation is evident in complex change categories, which require semantic understanding beyond visual recognition. 

Static prompts introduce semantic awareness through fixed text, providing consistent anchors to help the model associate visual patterns with semantic concepts. However, static prompts limit adaptability to diverse contexts and change types, resulting in limited improvement. Dynamic descriptions achieve context-aware text based on scene features and change patterns, enabling more relevant semantic guidance and enhancing alignment between visual features and semantics. LLM-enhanced reasoning leverages the reasoning capabilities of large language models to provide fine-grained semantic analysis and context. This enables the model to perform semantic reasoning, combining domain knowledge and logic to improve change detection accuracy and robustness.

\vspace{-10pt}
\paragraph{Bird's Eye View Module}
The BEV module addresses perspective distortion in aerial imagery, which can impact change detection accuracy, particularly for objects with varying heights and orientations. Without BEV correction, the model struggles with perspective-induced geometric inconsistencies, leading to false positives and missed changes. This is problematic in urban environments where building heights and viewing angles create perspective effects.

The results in Table.~\ref{table:bev} demonstrate that, simple geometric projection provides basic perspective normalization through fixed transformation matrices, offering initial correction for systematic perspective distortions. However, this approach lacks the flexibility to adapt to scene-specific geometric variations and complex topography. Learnable transformation introduces adaptive perspective correction through trainable parameters that adjust to scene-specific characteristics, enabling the model to learn optimal transformations for different scenes and improve spatial consistency.

Our multi-scale BEV integrates perspective information hierarchically across feature resolutions, enabling perspective correction that preserves both global geometric consistency and local spatial details. This ensures that perspective correction benefits are propagated throughout feature extraction, improving spatial consistency and edge preservation.


\section{Conclusion}

We propose SegChange-R1, a remote sensing change detection method leveraging LLMs for enhanced semantic reasoning. By integrating textual descriptions, our approach focuses on regions of interest, improving the accuracy and efficiency of change detection. To address modality misalignment, we designed a linear attention-based BEV converter, which aligns features into a unified spatial representation, enhancing spatial consistency and global context modeling. To support vision-language change detection, we introduce DVCD, a drone-view building change detection dataset featuring 13,800 image pairs across diverse urban and rural scenes. Experiments on four benchmark datasets demonstrate that SegChange-R1 achieves state-of-the-art performance in F1 score, IoU, and overall accuracy. Ablation studies validate the effectiveness of text-guided reasoning and the BEV module. This work opens new directions for integrating LLMs into remote sensing.

{\small
\bibliographystyle{ieee_fullname}
\bibliography{egbib}
}

\end{document}